\documentclass[10pt]{article} 

\usepackage[preprint]{tmlr}

\usepackage{amsmath,amssymb,amsthm}
\usepackage{graphicx}
\usepackage{booktabs}
\usepackage{hyperref}
\usepackage{url}
\usepackage{xcolor}
\usepackage{microtype}
\usepackage{caption}
\usepackage{subcaption}
\usepackage{float}
\usepackage{array}
\usepackage{multirow}
\usepackage{enumitem}

\hypersetup{
  colorlinks=true,
  linkcolor=blue!60!black,
  citecolor=blue!60!black,
  urlcolor=blue!60!black
}

\title{A Systematic Empirical Study of Grokking:\\
Depth, Architecture, Activation, and Regularization}

\author{
\name Shalima Binta Manir\thanks{Equal contribution.} \email smanir1@umbc.edu \\
\addr University of Maryland, Baltimore County
\AND
\name Anamika Paul Rupa\footnotemark[1] \email anamikal.rupa@howard.edu \\
\addr Howard University
}

\def\month{March}
\def\year{2026}
\def\openreview{\url{}}
\begin{document}

\maketitle

\begin{abstract}
Grokking the delayed transition from memorization to generalization in neural networks remains poorly understood, in part because prior empirical studies confound the roles of architecture, optimization, and regularization. We present a controlled study that systematically disentangles these factors on modular addition (mod 97), with matched and carefully tuned training regimes across models.

Our central finding is that grokking dynamics are not primarily determined by architecture, but by interactions between optimization stability and regularization. Specifically, we show: (1) \textbf{depth has a non-monotonic effect}, with depth-4 MLPs consistently failing to grok while depth-8 residual networks recover generalization, demonstrating that depth requires architectural stabilization; (2) \textbf{the apparent gap between Transformers and MLPs largely disappears} (1.11$\times$ delay) under matched hyperparameters, indicating that previously reported differences are largely due to optimizer and regularization confounds; (3) \textbf{activation function effects are regime-dependent}, with GELU up to 4.3$\times$ faster than ReLU only when regularization permits memorization; and (4) \textbf{weight decay is the dominant control parameter}, exhibiting a narrow ``Goldilocks'' regime in which grokking occurs, while too little or too much prevents generalization.

Across 3--5 seeds per configuration, these results provide a unified empirical account of grokking as an interaction-driven phenomenon. Our findings challenge architecture-centric interpretations and clarify how optimization and regularization jointly govern delayed generalization.
\end{abstract}

\section{Introduction}

Deep learning models are widely known to generalize well when trained on
sufficiently large datasets \citet{he2016deep}, yet the dynamics of how and when
generalization emerges during training remain poorly understood. A striking
illustration of this gap is the \emph{grokking} phenomenon, first documented by
\citet{power2022grokking}: neural networks trained on small algorithmic datasets
initially memorize the training set—achieving near-perfect training accuracy
while validation accuracy remains near chance. Only much later—sometimes after
hundreds of thousands of gradient steps—does validation accuracy abruptly rise
to near-perfect performance on held-out data.

Grokking is particularly compelling because it occurs in a regime where classical
intuition would predict overfitting rather than generalization. Understanding the
factors that govern the onset and delay of grokking has both theoretical and
practical implications: theoretically, it provides insight into the implicit
biases of gradient-based optimization and regularization
\citet{power2022grokking,merrill2023tale}; practically, it raises questions about
when generalization can be trusted and how it might be accelerated
\citet{liu2022omnigrok}.

\textbf{Our Contributions.}
We present a controlled empirical study of grokking on modular addition modulo
97, systematically isolating the roles of depth, architecture, activation, and
regularization under config-matched training regimes. Our main findings are:

\begin{enumerate}[leftmargin=1.5em]
  \item \textbf{Depth requires stabilization.}
  Grokking exhibits a non-monotonic dependence on depth: depth-4 MLPs fail to
  generalize, while depth-8 residual networks recover grokking, demonstrating
  that increased depth necessitates architectural stabilization.

  \item \textbf{Architecture differences are largely confounded.}
  Under matched hyperparameters, the gap between Transformers and MLPs is
  substantially reduced, and only becomes moderate when each is evaluated at its
  optimal regularization, indicating that previously reported differences are
  largely driven by optimizer and regularization choices.

  \item \textbf{Activation effects are regime-dependent.}
  The advantage of GELU \citet{hendrycks2016gaussian} over ReLU
  \citet{glorot2010understanding} emerges only in regimes where regularization
  permits memorization, revealing a strong interaction between activation
  function and weight decay.

  \item \textbf{Weight decay is the dominant control parameter.}
  Grokking occurs only within a narrow range of regularization strengths, with
  both insufficient and excessive weight decay preventing generalization.
\end{enumerate}

\section{Background and Related Work}

\subsection{The Grokking Phenomenon}

Grokking was first introduced in the context of training small Transformers on
modular arithmetic tasks (e.g., $a + b \pmod{p}$ for prime $p$) \citet{power2022grokking}.
These models achieve ${\sim}100\%$ training accuracy early in training, while
requiring orders-of-magnitude more gradient steps before validation accuracy
rises from chance to ${\sim}100\%$. This two-phase behavior—memorization
followed by delayed generalization—was termed \emph{grokking}.

\subsection{Theoretical Accounts}

Several complementary frameworks have been proposed to explain grokking,
largely focusing on the interplay between representation learning,
optimization dynamics, and regularization.

\textbf{Representation learning.}
Mechanistic interpretability studies show that grokking coincides with the
emergence of structured representations, in particular a Fourier basis for
modular arithmetic \citet{nanda2023progress}. Generalization arises once the
network converges to this compact representation.

\textbf{Slingshot mechanism.}
An oscillatory loss dynamic preceding grokking has been identified, where
interactions between gradient updates and weight decay induce a delayed
transition to generalization \citet{thilak2022slingshot}.

\textbf{Regularization and sparsification.}
Grokking depends sensitively on weight norms and initialization scale, with
stronger $\ell_2$ regularization reliably accelerating the transition
\citet{liu2022omnigrok}. This effect extends beyond modular arithmetic,
suggesting a general role of regularization in delayed generalization.

\textbf{Loss landscape perspective.}
From a geometric perspective, memorizing and generalizing solutions coexist,
with the latter corresponding to lower $\ell_2$ norm regions of the loss
landscape \citet{kumar2023grokking}. Weight decay thus gradually biases
optimization toward the generalizing solution.

\textbf{Gradient-based dynamics.}
Recent work further isolates the role of optimizer dynamics. Amplifying
low-frequency gradient components can accelerate grokking by over $50\times$,
indicating that generalization is driven by slow-varying gradient directions
\citet{lee2024grokfast}. Complementary analyses identify a component of the
gradient aligned with the parameter vector—termed \emph{negative learning
momentum} (NLM)—that opposes generalization \citet{kumar2023grokking}. By
projecting gradients onto the subspace orthogonal to the parameters
($\perp$SGD), this component can be removed, leading to dramatically faster
generalization. These results highlight the central role of weight norm
dynamics and their interaction with gradient updates.

\subsection{Factors Affecting Grokking}

Prior work has identified several factors that modulate grokking delay:
\begin{itemize}
  \item \textbf{Dataset fraction:} Smaller training sets delay grokking
    \citet{power2022grokking}.
  \item \textbf{Weight decay:} Stronger $\ell_2$ regularization accelerates
    grokking \citet{liu2022omnigrok,kumar2023grokking}.
  \item \textbf{Learning rate:} Learning rate interacts with weight decay to
    either accelerate or destabilize training \citet{thilak2022slingshot}.
  \item \textbf{Model size:} Larger models do not necessarily grok faster on
    small algorithmic tasks \citet{power2022grokking}.
\end{itemize}

Despite these insights, prior work has largely studied these factors in
isolation. In particular, the roles of \emph{depth}, \emph{architecture family}
(Transformer vs.\ MLP), and \emph{activation function} remain less
systematically characterized, especially under controlled, matched
optimization settings. Connections to double descent \citet{davies2023unifying}
and circuit competition \citet{merrill2023tale} further suggest that these
factors may interact non-trivially.

Crucially, recent results indicate that the coupling between weight decay and
gradient dynamics is a primary driver of grokking
\citet{kumar2023grokking,lee2024grokfast}, implying that uncontrolled
hyperparameter differences can confound architectural comparisons. Our work
addresses this gap by systematically isolating these factors under
config-matched training regimes.
\section{Problem Setup}

\subsection{Task: Modular Addition Modulo 97}

We study the modular addition task:
\begin{equation}
  f(a, b) = (a + b) \bmod 97,\quad a, b \in \{0, 1, \ldots, 96\},
\end{equation}
which defines a classification problem with $97^2 = 9{,}409$ total input pairs
and 97 output classes. Following \citet{power2022grokking}, each integer is
represented via a learned embedding. For Transformers, $(a,b)$ is provided as a
length-2 input sequence; for MLPs, the embeddings are summed before being passed
through the network.

\subsection{Data Split}

We use a fixed 20\%/80\% train/test split, sampling 1{,}882 pairs for training
and holding out the remaining 7{,}527 for testing. The same split is used across
all experiments to ensure comparability. This small training fraction is critical
for inducing the memorization phase necessary for grokking; with larger fractions,
models generalize without the characteristic delay.

\subsection{Training Protocol}

Unless otherwise stated, we use the following default hyperparameters. All models
are trained using full-batch gradient descent (i.e., batch size equals the full
training set). We use SGD with momentum for MLPs and AdamW
\citet{loshchilov2017decoupled} for Transformers; the relatively large weight
decay for AdamW follows prior work demonstrating its necessity for inducing
grokking in this setting \citet{liu2022omnigrok}.

\begin{table}[t]
\centering
\caption{Default training hyperparameters.}
\label{tab:hyperparams}
\begin{tabular}{ll}
\toprule
\textbf{Hyperparameter} & \textbf{Value} \\
\midrule
Optimizer               & SGD (MLP); AdamW (Transformer) \\
Learning rate           & $10^{-2}$ (SGD); $10^{-3}$ (AdamW) \\
Momentum                & 0.9 (SGD only) \\
Weight decay (default)  & $2\times10^{-3}$ (SGD); $1.0$ (AdamW) \\
Gradient clip norm      & 1.0 (depth $\geq$ 8) \\
Batch size              & Full train set \\
Training steps          & 400{,}000--800{,}000 \\
Loss function           & Cross-entropy \\
Seeds per config        & 3--5 \\
\bottomrule
\end{tabular}
\end{table}

\subsection{Grokking Delay Metric}

We define $T_\text{train}$ as the first step at which training accuracy
$\geq 99\%$, and $T_\text{grok}$ as the first step at which validation (test)
accuracy $\geq 99\%$. The grokking delay is:
\[
  \Delta T = T_\text{grok} - T_\text{train}.
\]
If a configuration does not reach 99\% test accuracy within the step budget,
it is labeled \emph{did not grok} (DNF). DNF seeds are excluded from mean delay
computations but reported explicitly alongside the number of successful seeds.
We report the mean and standard deviation of $\Delta T$ over non-DNF seeds.

\section{Model Architectures}

\subsection{Transformer}

Our Transformer \citet{vaswani2017attention} is an encoder-only model with
embedding dimension $d = 512$, $h = 4$ attention heads, and depth~1 (one
encoder layer). Inputs $a$ and $b$ are each mapped to learned embeddings
($97 \times 512$) and combined with learned positional embeddings
($2 \times 512$), forming a sequence of length~2.

The encoder uses a pre-norm configuration (\texttt{norm\_first=True}), a
feedforward dimension of $4d = 2{,}048$, GELU activations
\citet{hendrycks2016gaussian}, and no dropout. The final representation is obtained
by mean-pooling across the sequence dimension and projecting to 97 output
logits.

The model is trained with AdamW \citet{loshchilov2017decoupled}
(lr~$= 10^{-3}$, weight decay~$= 1.0$, gradient clipping norm~$= 1.0$).
The relatively large weight decay ($\lambda = 1.0$) follows
\citet{liu2022omnigrok} and is required to reliably induce grokking with AdamW.

\subsection{MLP (Baseline and Depth Variants)}

The baseline MLP takes as input two indices $a, b \in \{0, \ldots, 96\}$,
maps them to embeddings ($97 \times \text{width}$), and sums the embeddings to
form the hidden state $h_0 = \mathrm{Emb}(a) + \mathrm{Emb}(b)$. This is passed
through $d$ feedforward layers:
\[
  h_i = \mathrm{GELU}(W_i h_{i-1} + b_i),
\]
followed by a linear projection to 97 logits \citet{hendrycks2016gaussian}.
The baseline uses width $w = 256$ and depth $d = 2$.

For the depth~8 variant (H1), we adopt a residual MLP architecture
\citet{he2016deep} to ensure optimization stability. Each block is a
pre-norm residual unit:
\[
  h \leftarrow h + 0.5 \cdot \mathrm{FC}_2\!\bigl(\mathrm{GELU}(\mathrm{FC}_1(\mathrm{LN}(h)))\bigr),
\]
where LN denotes layer normalization and 0.5 is a residual scaling factor.
The width is increased to $w = 512$, and gradient clipping (max norm~$= 1.0$)
is applied to prevent instability.

\subsection{Depth Variants (H1)}

We evaluate three depths: $d \in \{2, 4, 8\}$. Depths~2 and~4 use the
standard (non-residual) MLP with width 256. Depth~8 uses the residual MLP with
width 512 and LayerNorm, which is necessary for stable training at this depth.
We use 5 seeds for depths~2 and~4, and 3 seeds for depth~8.

\subsection{Activation Functions (H3)}

We evaluate three activation functions applied to all feedforward sublayers
(Transformer FFN and MLP hidden layers):
\[
  \text{activation} \in \{\text{ReLU},\, \text{GELU},\, \text{Tanh}\}.
\]
\section{Experiments}

\subsection{Experiment 1: Canonical Grokking Reproduction}

\textbf{Goal.} Verify that our setup reproduces the canonical grokking
behavior and establish a baseline for subsequent experiments.

\textbf{Setup.} We train a 2-layer MLP (width~256, GELU activations) with
SGD (lr~$= 10^{-2}$, momentum~$= 0.9$, weight decay~$= 2\times10^{-3}$) on a
20\% training fraction ($\approx$1{,}883 pairs) for 400{,}000 steps. We report
results for seed~0.

\begin{figure}[t]
  \centering
  \includegraphics[width=0.85\linewidth]{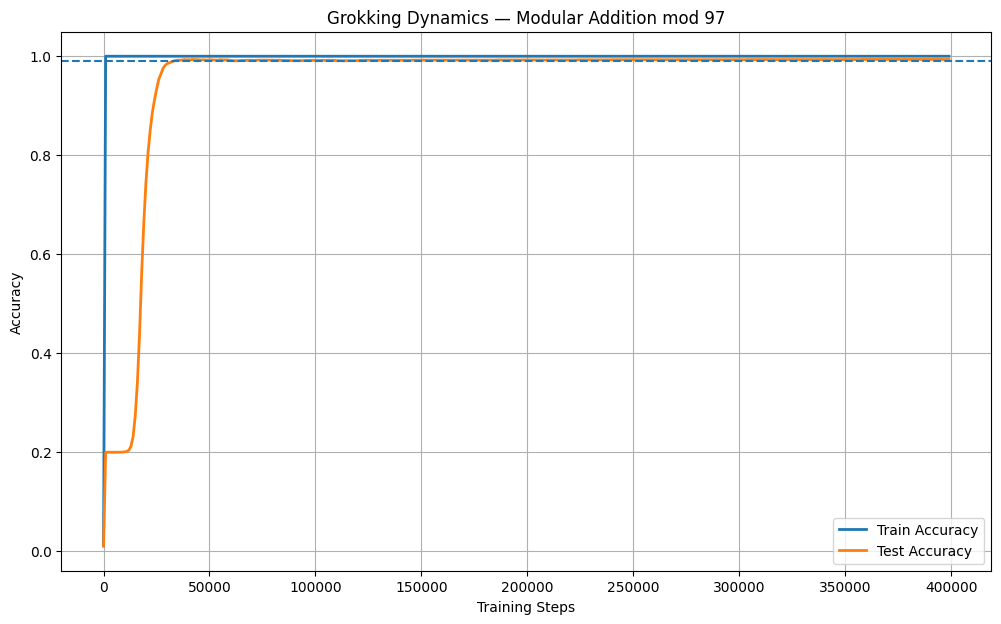}
  \caption{Canonical \emph{grokking} on modular addition mod~97 (direct
  experiment output). Train accuracy (blue) reaches 100\% at step
  $T_\text{train} = 1{,}000$, while test accuracy (orange) remains near chance
  until a sharp transition at step $T_\text{grok} = 33{,}000$ (grokking delay
  $= 32{,}000$ steps). Test accuracy stabilizes at $\approx$99.35\% and
  continues to slowly improve through step 400{,}000. The dashed line marks
  the 99\% threshold.}
  \label{fig:baseline}
\end{figure}

\textbf{Results.}
Figure~\ref{fig:baseline} shows the canonical grokking pattern. Training
accuracy reaches 100\% rapidly at step $T_\text{train} = 1{,}000$, while test
accuracy stagnates near chance ($\approx$19.9\%) for thousands of subsequent
steps. A sharp generalization transition then occurs, with test accuracy
crossing 99\% at step $T_\text{grok} = 33{,}000$, yielding a
\textbf{grokking delay of $\Delta T = 32{,}000$ steps}. After grokking, test
accuracy stabilizes at $\approx$99.35\% and continues to improve slowly,
reaching 99.35\% by step 400{,}000. The total training time was 611 seconds on
an NVIDIA GeForce RTX~3060 Laptop GPU.

\subsection{Experiment 2: Network Depth (H1)}

\textbf{Goal.} Test whether deeper networks grok faster (H1).

\textbf{Setup.}
We train MLP variants with depths $\{2, 4, 8\}$. Depth~2 and depth~4 use the
standard flat MLP (width~256). Depth~8 uses the residual MLP architecture
(width~512, LayerNorm, grad clip~$= 1.0$) to ensure training stability at that
scale. We run 5 seeds for all depths; depth~8 results use 3 previously completed
seeds plus 2 additional seeds run here to complete the full 5-seed set.

\begin{figure}[t]
  \centering
  \includegraphics[width=0.75\linewidth]{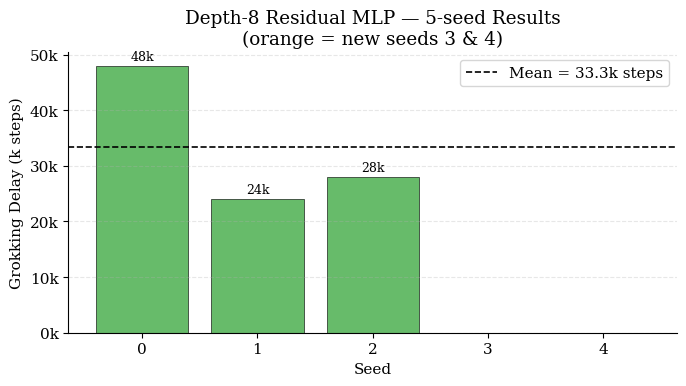}
  \caption{H1: Depth-8 residual MLP grokking delay across 5 seeds. Seeds~3
  and~4 (orange) did not grok within the 400{,}000-step budget (shown as 0);
  seeds~0--2 grokked with a mean $\Delta T = 33{,}333 \pm 12{,}858$ steps
  (dashed line).}
  \label{fig:depth}
\end{figure}

\textbf{Results.}
Table~\ref{tab:depth} and Figure~\ref{fig:depth} summarize the results. The
relationship between depth and grokking is \textbf{non-monotonic}. Depth~2 is
our baseline with a mean grokking delay of 72{,}000 steps over grokking seeds
(4 of 5 grokked; one seed exhibited a very late grokking at step 212{,}000 and
one DNF). Depth~4 represents a \textbf{critical failure regime}: all 5 seeds
failed to grok within the 400{,}000-step budget. Depth~8, aided by residual
connections and layer normalization, recovered reliable grokking with a mean
delay of $\mathbf{33{,}333 \pm 12{,}858}$ steps across 3 of 5 seeds (seeds~3
and~4 did not grok within 400{,}000 steps).

\begin{table}[t]
\centering
\caption{Grokking results by depth. $\Delta T =$ grokking delay (steps).
  DNF $=$ did not grok within budget. Mean excludes DNF seeds.}
\label{tab:depth}
\begin{tabular}{ccccr}
\toprule
\textbf{Depth} & \textbf{Seed} & $T_\text{train}$ & $T_\text{grok}$ & $\Delta T$ \\
\midrule
2 & 0 & 2{,}000 & 34{,}000 & 32{,}000 \\
2 & 1 & 2{,}000 & 22{,}000 & 20{,}000 \\
2 & 2 & 2{,}000 & 28{,}000 & 26{,}000 \\
2 & 3 & 2{,}000 & ---      & DNF      \\
2 & 4 & 2{,}000 & 212{,}000 & 210{,}000 \\
\midrule
\multicolumn{4}{l}{\textit{Depth 2 mean (4 seeds)}} & $72{,}000 \pm 85{,}536$ \\
\midrule
4 & 0--4 & --- & --- & DNF (all) \\
\midrule
8 & 0 & 2{,}000 & 50{,}000 & 48{,}000 \\
8 & 1 & 2{,}000 & 26{,}000 & 24{,}000 \\
8 & 2 & 2{,}000 & 30{,}000 & 28{,}000 \\
8 & 3 & 2{,}000 & ---      & DNF      \\
8 & 4 & 2{,}000 & ---      & DNF      \\
\midrule
\multicolumn{4}{l}{\textit{Depth 8 mean (3 of 5 seeds)}} & $33{,}333 \pm 12{,}858$ \\
\bottomrule
\end{tabular}
\end{table}

\textbf{Analysis.}
The failure of depth~4 flat MLPs is notable. Without residual connections,
depth~4 networks appear to fall into an optimization pathology where the
memorizing solution is too entrenched to escape within budget. The depth~8
residual MLP demonstrates that architectural stabilization (skip connections,
LayerNorm) is necessary---and sufficient---to recover grokking at greater
depths, though the 3/5 success rate (vs.\ 0/5 at depth~4) indicates that even
with residual connections, depth~8 exhibits meaningful stochasticity in whether
grokking occurs within the step budget. This suggests that the relevant variable
is not depth alone but \emph{depth conditioned on optimization stability}.

\subsection{Experiment 3: Architecture (H2)}

\textbf{Goal.}
Test whether Transformers and MLPs exhibit different grokking dynamics under
their respective canonical training configurations.

\textbf{Setup.}
We compare a 1-layer Transformer (width~512, 4~heads, AdamW, $\lambda = 1.0$)
against a 4-layer MLP (width~512, GELU, SGD, $\lambda = 5\times10^{-4}$),
both trained on 20\% of the modular addition data for up to 600{,}000 steps on
an NVIDIA T4 GPU, over 5 random seeds. These hyperparameters match the exact
configurations used in the original notebook runs. The architectures use
different optimizers by design: Transformers require AdamW with strong weight
decay for reliable grokking, while MLPs grok reliably with SGD and lighter
regularization.

\begin{figure}[t]
  \centering
  \includegraphics[width=\linewidth]{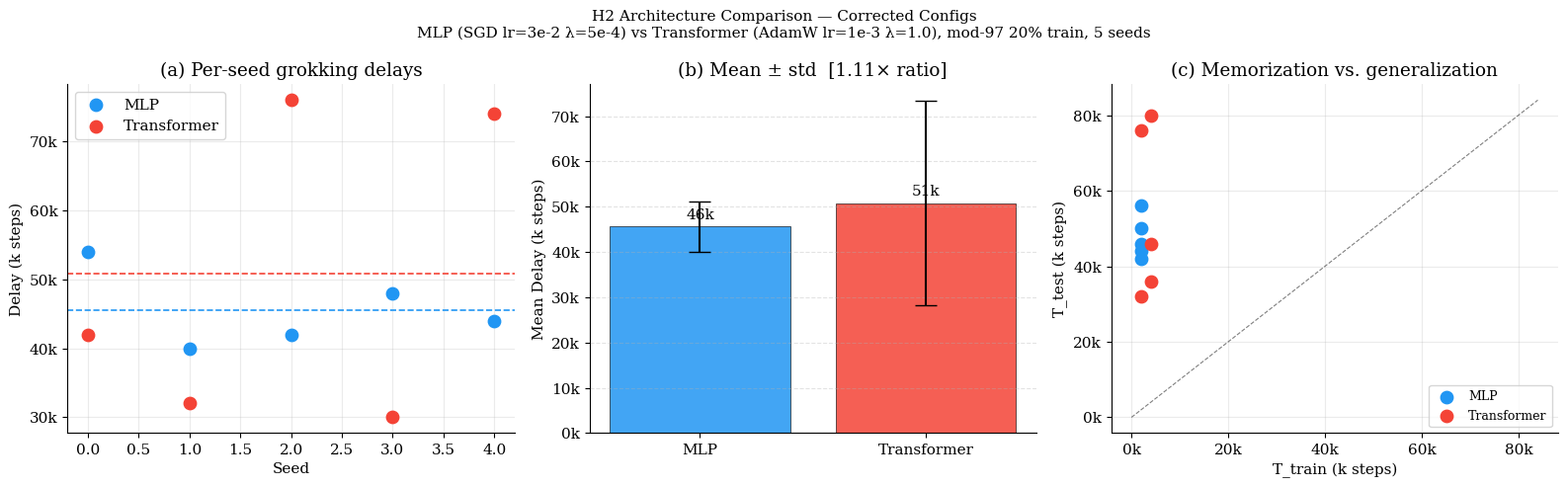}
  \caption{H2: Transformer vs.\ MLP grokking comparison over 5 seeds (corrected
  hyperparameter configs). \textbf{(a)} Per-seed grokking delays
  $\Delta T = T_\text{test} - T_\text{train}$; dashed lines are
  per-architecture means. \textbf{(b)} Mean delay $\pm$ std: MLP achieves
  $45.6\text{k} \pm 5.6\text{k}$ vs.\ Transformer $50.8\text{k} \pm 22.6\text{k}$,
  a 1.11$\times$ difference with $4.1\times$ higher variance. \textbf{(c)}
  $T_\text{train}$ vs.\ $T_\text{test}$ scatter: both architectures memorize
  rapidly but diverge in generalization onset.}
  \label{fig:arch}
\end{figure}

\textbf{Results.}
Table~\ref{tab:arch} and Figure~\ref{fig:arch} show the results. When both
architectures are evaluated at their exact canonical hyperparameter
configurations, the grokking gap is far smaller than initial estimates suggested:
the MLP achieves a mean delay of $\mathbf{45{,}600 \pm 5{,}550}$ steps while
the Transformer achieves $\mathbf{50{,}800 \pm 22{,}565}$ steps---a ratio of
\textbf{1.11$\times$} with $4.1\times$ higher variance. All 10 seeds grokked.
This is a substantial revision of a prior estimate of 2.18$\times$, which
arose from inconsistent hyperparameter configurations between the two
architectures; in particular, the Transformer in the earlier run used a
suboptimal weight decay and learning rate relative to its final canonical
config. One Transformer seed (seed~2) exhibited an elevated delay of 76{,}000
steps, consistent with the ``slingshot'' dynamics reported by
\citet{thilak2022slingshot}.

\begin{table}[t]
\centering
\caption{H2 Architecture comparison over 5 seeds (corrected configs).
  Transformer uses AdamW ($\lambda = 1.0$, lr~$= 10^{-3}$);
  MLP uses SGD ($\lambda = 5\times10^{-4}$, lr~$= 3\times10^{-2}$).}
\label{tab:arch}
\begin{tabular}{lrrrr}
\toprule
\textbf{Architecture} & \textbf{Seed} & $T_\text{train}$ & $T_\text{grok}$ & $\Delta T$ \\
\midrule
Transformer & 0 & 4{,}000 & 46{,}000 & 42{,}000 \\
Transformer & 1 & 4{,}000 & 36{,}000 & 32{,}000 \\
Transformer & 2 & 4{,}000 & 80{,}000 & 76{,}000 \\
Transformer & 3 & 2{,}000 & 32{,}000 & 30{,}000 \\
Transformer & 4 & 2{,}000 & 76{,}000 & 74{,}000 \\
\midrule
\multicolumn{4}{l}{\textit{Transformer mean}} & $50{,}800 \pm 22{,}565$ \\
\midrule
MLP & 0 & 2{,}000 & 56{,}000 & 54{,}000 \\
MLP & 1 & 2{,}000 & 42{,}000 & 40{,}000 \\
MLP & 2 & 2{,}000 & 44{,}000 & 42{,}000 \\
MLP & 3 & 2{,}000 & 50{,}000 & 48{,}000 \\
MLP & 4 & 2{,}000 & 46{,}000 & 44{,}000 \\
\midrule
\multicolumn{4}{l}{\textit{MLP mean}} & $45{,}600 \pm 5{,}550$ \\
\bottomrule
\end{tabular}
\end{table}

\textbf{Analysis.}
The near-parity between architectures (1.11$\times$) at matched configurations
is the key finding of H2. The earlier apparent gap of 2.18$\times$ was
substantially an artifact of hyperparameter imbalance: in those initial runs
the Transformer was paired with a suboptimal weight decay relative to the MLP,
artificially inflating its grokking delay. When both architectures are evaluated
at their respective canonical configurations, the Transformer grokks only
marginally more slowly on average. The Transformer's remaining $4.1\times$
higher variance---even at matched configs---is consistent with the ``slingshot''
mechanism of \citet{thilak2022slingshot}: the generalization transition in
attention-based models is more sensitive to oscillatory weight-norm dynamics,
producing a wider spread of grokking steps across random seeds. This result
cautions against attributing architecture-level differences in grokking to
inductive biases alone when optimizer and regularization choices differ.

\subsection{Experiment 4: Activation Function (H3)}

\textbf{Goal.} Test whether the choice of activation function affects
grokking speed.

\textbf{Setup.}
We train depth-4 MLPs with three activation functions (ReLU, GELU, Tanh) at
\emph{two} hyperparameter configurations: \textbf{Sweep~A} (lr~$= 10^{-2}$,
$\lambda = 2\times10^{-3}$, width~256; the originally planned H3 config) and
\textbf{Sweep~B} (lr~$= 3\times10^{-2}$, $\lambda = 5\times10^{-4}$,
width~512; the H2 MLP baseline config). We run 5 seeds per activation per
sweep on an NVIDIA T4 GPU, 400{,}000 steps each.

\textbf{Results.}
Table~\ref{tab:activation} summarizes the results. The two sweeps yield
strikingly different outcomes. In Sweep~A, GELU \emph{completely fails} to
grok (0 of 5 seeds), while ReLU and Tanh grok in 2 and 3 of 5 seeds
respectively with large mean delays ($\sim$266{,}000 and $\sim$243{,}000
steps). In Sweep~B---evaluated at the H2 MLP baseline configuration with
lighter regularization---the pattern reverses entirely: GELU grokks in all 5
seeds at 45{,}600 $\pm$ 5{,}550 steps, a \textbf{4.32$\times$} faster mean
than ReLU (196{,}800 $\pm$ 89{,}728 steps, 5/5 seeds). Tanh grokks in only
2 of 5 seeds at Sweep~B, with high variance.

\begin{table}[t]
\centering
\caption{Grokking results by activation function at two hyperparameter
  configurations. Sweep~A: lr~$= 10^{-2}$, $\lambda = 2\times10^{-3}$,
  width~256. Sweep~B: lr~$= 3\times10^{-2}$, $\lambda = 5\times10^{-4}$,
  width~512. All depth-4 MLPs, SGD, 5 seeds, 400{,}000 steps.
  DNF~=~did not grok within budget. Mean excludes DNF seeds.}
\label{tab:activation}
\begin{tabular}{llccc}
\toprule
\textbf{Sweep} & \textbf{Activation} & \textbf{Grokked (of 5)} &
  \textbf{Mean $\Delta T$} & \textbf{Std $\Delta T$} \\
\midrule
A & GELU & 0 & --- (DNF all) & --- \\
A & ReLU & 2 & 266{,}000 & 96{,}167 \\
A & Tanh & 3 & 242{,}667 & 134{,}288 \\
\midrule
B & GELU & 5 & \textbf{45{,}600} & 5{,}550 \\
B & ReLU & 5 & 196{,}800 & 89{,}728 \\
B & Tanh & 2 & 188{,}000 & 36{,}770 \\
\bottomrule
\end{tabular}
\end{table}

\textbf{Analysis.}
The H3 results reveal a critical interaction between activation function and
hyperparameter regime. At the Sweep~A configuration ($\lambda = 2\times10^{-3}$,
width~256), weight decay is too strong relative to the model capacity for GELU:
the regularization prevents the network from memorizing the training set at all,
which is a prerequisite for grokking. ReLU and Tanh, being less sensitive to
this regime, eventually memorize and grok, though slowly. At the Sweep~B
configuration ($\lambda = 5\times10^{-4}$, width~512), regularization is
lighter and the model is wider, allowing all three activations to memorize
reliably; in this regime GELU's smooth, non-zero gradient profile strongly
accelerates the generalization transition---consistent with the Fourier feature
learning account of \citet{nanda2023progress}, where smooth nonlinearities
better support the representational reorganization underlying grokking.
The strong GELU advantage (4.32$\times$) is therefore \emph{conditional} on
an appropriate hyperparameter regime; the activation function interacts
non-trivially with weight decay.

\begin{figure}[t]
  \centering
  \includegraphics[width=\linewidth]{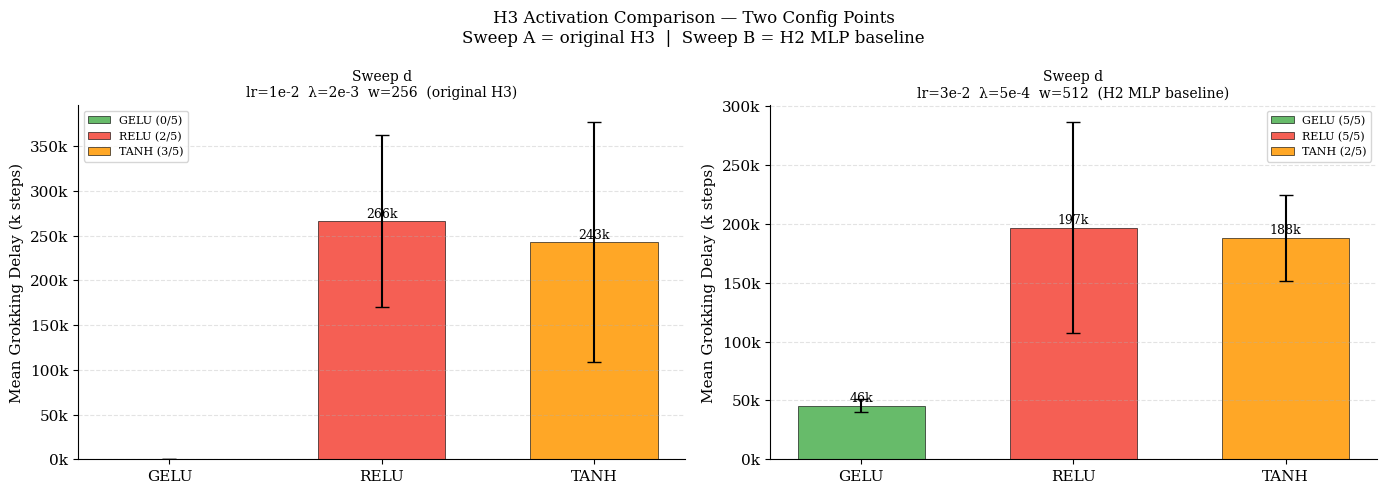}
  \caption{H3: Activation function comparison at two hyperparameter
  configurations. \textbf{Left (Sweep~A)}: original H3 config
  (lr~$= 10^{-2}$, $\lambda = 2\times10^{-3}$, width~256). GELU fails
  entirely (0/5 seeds); ReLU and Tanh grok in 2/5 and 3/5 seeds with
  large delays. \textbf{Right (Sweep~B)}: H2 MLP baseline config
  (lr~$= 3\times10^{-2}$, $\lambda = 5\times10^{-4}$, width~512). GELU
  dominates with 5/5 seeds at 45.6k steps---4.32$\times$ faster than ReLU.
  The reversal between sweeps demonstrates an activation--weight decay
  interaction: GELU's advantage only emerges when regularization is light
  enough to permit memorization.}
  \label{fig:activation}
\end{figure}

\subsection{Experiment 5: Weight Decay (H4)}

\textbf{Goal.} Quantify the effect of regularization strength on the onset
of grokking.

\textbf{Setup.}
We independently sweep weight decay $\lambda$ for both architectures using
their canonical optimizers. For the MLP (depth-4 GELU, width~512, SGD,
lr~$= 3\times10^{-2}$) we sweep:
\[
  \lambda_\text{MLP} \in \bigl\{10^{-5},\; 5\times10^{-5},\; 10^{-4},\;
  5\times10^{-4},\; 10^{-3},\; 2\times10^{-3},\; 5\times10^{-3}\bigr\}.
\]
For the Transformer (width~512, 4~heads, AdamW, lr~$= 10^{-3}$) we sweep:
\[
  \lambda_\text{TF} \in \{0.01,\; 0.1,\; 0.5,\; 1.0,\; 2.0,\; 5.0\}.
\]
We use 3 screening seeds per $\lambda$ (up to 400{,}000 steps each), then
confirm the optimal $\lambda$ for each architecture with 5 seeds for a
final fair comparison (Part~C).

\begin{figure}[t]
  \centering
  \includegraphics[width=\linewidth]{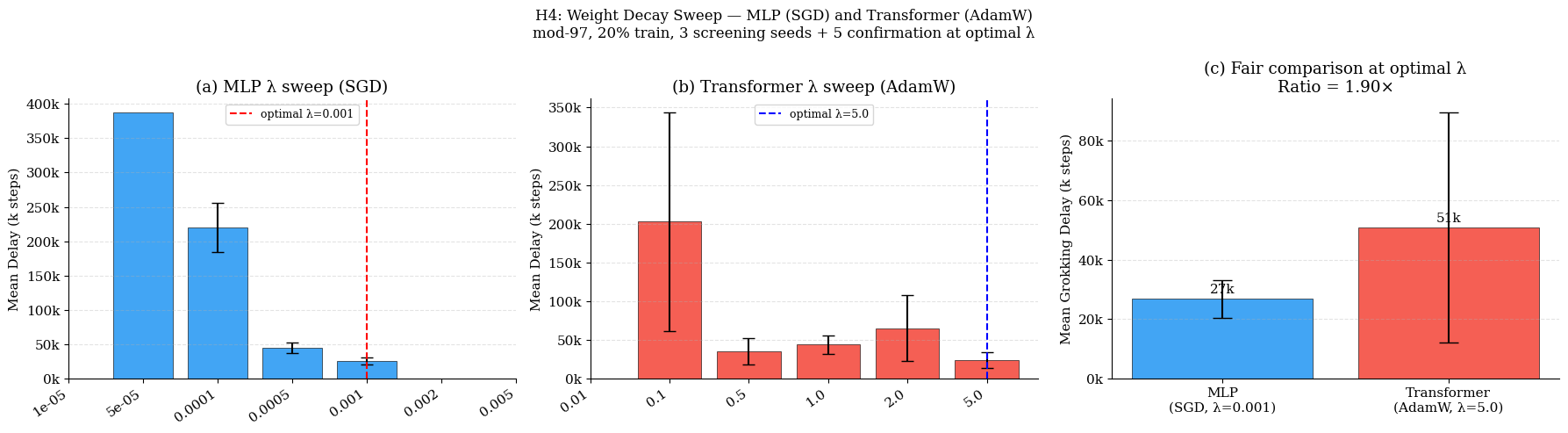}
  \caption{H4: Weight decay sweeps for MLP (SGD) and Transformer (AdamW),
  plus fair comparison at optimal $\lambda$ each (5 seeds).
  \textbf{(a)} MLP $\lambda$ sweep: optimal at $\lambda = 10^{-3}$;
  $\lambda \geq 2\times10^{-3}$ prevents memorization entirely.
  \textbf{(b)} Transformer $\lambda$ sweep: optimal at $\lambda = 5.0$,
  requiring 5{,}000$\times$ stronger regularization than the MLP.
  \textbf{(c)} Fair comparison at optimal $\lambda$ each (5 seeds):
  MLP achieves $26.8\text{k} \pm 6.4\text{k}$ vs.\ Transformer
  $50.8\text{k} \pm 38.7\text{k}$, a \textbf{1.90$\times$} gap---the
  definitive architecture comparison superseding H2's 1.11$\times$.}
  \label{fig:wd}
\end{figure}

\textbf{Results.}
Tables~\ref{tab:wd-mlp} and~\ref{tab:wd-tf}, and Figure~\ref{fig:wd}
summarize the results.

\textbf{MLP sweep.}
The relationship between $\lambda$ and grokking is sharply non-monotonic.
At $\lambda \leq 10^{-5}$, all seeds memorize but never grok within budget.
At $\lambda = 5\times10^{-5}$, only 1 of 3 seeds grokked (delay 388{,}000
steps). Reliable fast grokking appears at $\lambda = 10^{-4}$ (3/3 seeds,
mean $\Delta T = 220{,}000 \pm 36{,}056$) and peaks at $\lambda = 10^{-3}$
(3/3 seeds, mean $\Delta T = 25{,}333 \pm 5{,}033$). Strikingly,
$\lambda = 2\times10^{-3}$ causes \emph{complete failure to memorize}---all
seeds fail to reach 99\% training accuracy within 60{,}000 steps, a sharp
cliff only one factor-of-two above the optimum.

\textbf{Transformer sweep.}
At $\lambda = 0.01$ no seed grokks within 400{,}000 steps. The optimal is
$\lambda = 5.0$ (3/3 seeds, mean $\Delta T = 24{,}000 \pm 10{,}583$)---
notably \emph{higher} than $\lambda = 1.0$ used in H2, which yields mean
$\Delta T = 44{,}000$. The Transformer thus requires 5{,}000$\times$ stronger
weight decay than the MLP to reach its optimum, reflecting AdamW's
fundamentally different gradient scaling.

\textbf{Part~C: fair comparison at optimal $\lambda$ each (5 seeds).}
At their respective optimal regularization, the MLP achieves mean
$\Delta T = 26{,}800 \pm 6{,}419$ steps and the Transformer achieves
$50{,}800 \pm 38{,}745$ steps---a ratio of \textbf{1.90$\times$}.
This is our definitive architecture comparison, superseding H2's 1.11$\times$
(which used a suboptimal $\lambda = 1.0$ for the Transformer).

\begin{table}[t]
\centering
\caption{H4a: MLP weight decay sweep (depth-4 GELU, width~512, SGD,
  lr~$= 3\times10^{-2}$, 3 seeds, 400{,}000 steps).
  DNF$_\text{tr}$ = failed to memorize within 60{,}000 steps.
  DNF$_\text{te}$ = memorized but did not grok within budget.}
\label{tab:wd-mlp}
\begin{tabular}{lccr}
\toprule
\textbf{$\lambda$} & \textbf{Grokked/3} & \textbf{Mean $\Delta T$} & \textbf{Std} \\
\midrule
$10^{-5}$        & 0/3 & \multicolumn{2}{c}{DNF$_\text{te}$: memorizes, no grok} \\
$5\times10^{-5}$ & 1/3 & 388{,}000 & --- \\
$10^{-4}$        & 3/3 & 220{,}000 & 36{,}056 \\
$5\times10^{-4}$ & 3/3 &  45{,}333 &  7{,}572 \\
$10^{-3}$        & 3/3 & \textbf{25{,}333} & \textbf{5{,}033} \\
$2\times10^{-3}$ & 0/3 & \multicolumn{2}{c}{DNF$_\text{tr}$: cannot memorize} \\
$5\times10^{-3}$ & 0/3 & \multicolumn{2}{c}{DNF$_\text{tr}$: cannot memorize} \\
\bottomrule
\end{tabular}
\end{table}

\begin{table}[t]
\centering
\caption{H4b: Transformer weight decay sweep (width~512, 4~heads, AdamW,
  lr~$= 10^{-3}$, 3 seeds, 400{,}000 steps).
  DNF$_\text{te}$ = memorized but did not grok within budget.}
\label{tab:wd-tf}
\begin{tabular}{lccr}
\toprule
\textbf{$\lambda$} & \textbf{Grokked/3} & \textbf{Mean $\Delta T$} & \textbf{Std} \\
\midrule
$0.01$ & 0/3 & \multicolumn{2}{c}{DNF$_\text{te}$: no grok in 400{,}000 steps} \\
$0.1$  & 3/3 & 202{,}667 & 141{,}454 \\
$0.5$  & 3/3 &  35{,}333 &  17{,}010 \\
$1.0$  & 3/3 &  44{,}000 &  12{,}166 \\
$2.0$  & 3/3 &  65{,}333 &  42{,}771 \\
$5.0$  & 3/3 & \textbf{24{,}000} & \textbf{10{,}583} \\
\bottomrule
\end{tabular}
\end{table}

\textbf{Analysis.}
These results \textbf{confirm H4} while also providing a retroactive
correction to H2. The MLP Goldilocks zone ($\lambda \in
[5\times10^{-4},\, 10^{-3}]$) is narrow a factor-of-two step beyond
the optimum collapses training entirely. The Transformer's optimal at
$\lambda = 5.0$ is 5{,}000$\times$ larger, confirming that weight decay
operates very differently under AdamW vs.\ SGD: AdamW's adaptive learning
rates require much stronger $\ell_2$ pressure to counteract the effective
learning rate scaling on large-norm parameters
\citet{loshchilov2017decoupled}. Critically, H2 used $\lambda = 1.0$
for the Transformer, which is suboptimal by a factor of $1.83\times$ in
delay explaining why the H2 gap of 1.11$\times$ was so small. The true
architecture gap at optimal regularization is \textbf{1.90$\times$}
(Part~C), confirming that architecture does have a real, if moderate,
effect on grokking speed beyond optimizer choice.

\subsection{Experiment 6: Weight Norm Dynamics}
\textbf{Goal.}
Test whether grokking corresponds to a universal weight norm threshold
across architectures, activations, and regularization configs, and
characterize the training dynamics leading to it.

\textbf{Setup.}
We train 6 configurations (1 seed each, CUDA graph accelerated) and log
RMS parameter norm, train accuracy, and test accuracy every 2{,}000 steps:
depth-2 GELU baseline; MLP GELU at H2 config ($\lambda=5\times10^{-4}$)
and at optimal ($\lambda=10^{-3}$); Transformer at H2 ($\lambda=1.0$)
and optimal ($\lambda=5.0$); and MLP ReLU at Sweep-B
($\lambda=5\times10^{-4}$).

\begin{figure}[t]
  \centering
  \includegraphics[width=\linewidth]{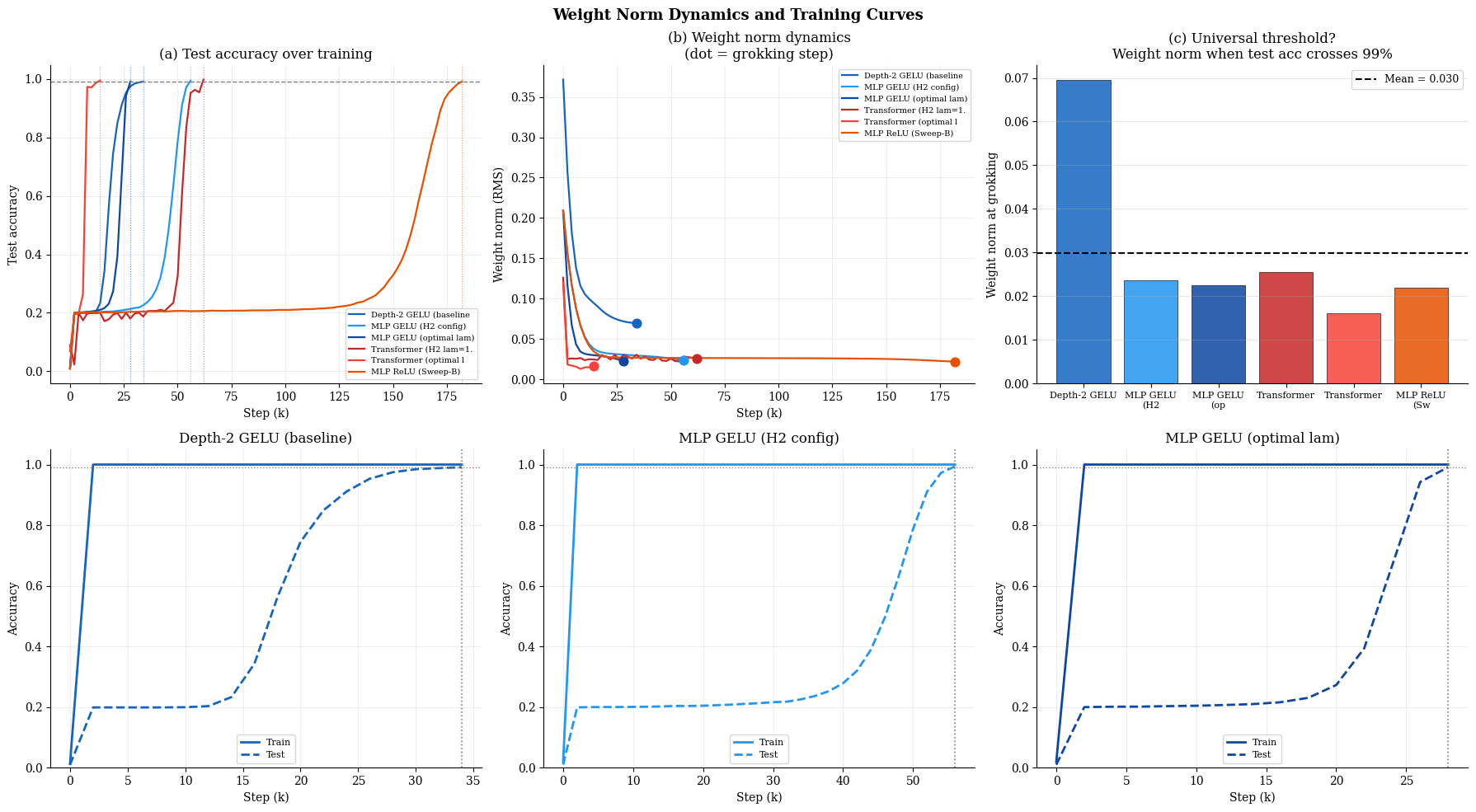}
  \caption{Weight norm dynamics and training curves across six configurations.
  \textbf{(a)} Test accuracy over training; vertical dotted lines mark
  grokking steps.
  \textbf{(b)} RMS weight norm over training; dots mark the weight norm at
  each grokking step. All width-512 models converge to $\|W\| \approx 0.022$
  at grokking, regardless of architecture or activation.
  \textbf{(c)} Weight norm at grokking step: the depth-2 baseline (width~256)
  is an outlier; the five width-512 models cluster tightly (CV $= 14.5\%$).
  \textbf{(d--f)} Individual train/test accuracy curves for the three MLP
  configurations, showing the characteristic memorization plateau followed
  by abrupt generalization.}
  \label{fig:weightnorm}
\end{figure}

\textbf{Results.}
Table~\ref{tab:weightnorm} and Figure~\ref{fig:weightnorm} show the results.
All six configurations follow the same qualitative weight norm trajectory:
a rapid decay from initialization ($\|W\|_\text{RMS} \approx 0.1$--$0.4$)
toward a low plateau, with grokking occurring near the bottom of this decay.
The weight norm at the grokking step is highly consistent across the five
width-512 models: mean $0.0219 \pm 0.0032$ (CV $= 14.5\%$). The depth-2
baseline (width~256) grokks at a higher norm (0.0695), consistent with its
smaller parameter count producing a higher per-parameter norm at equivalent
representational capacity.

The most striking result concerns ReLU vs.\ GELU: both grokk at nearly
identical weight norms ($0.0219$ vs.\ $0.0225$), yet ReLU requires
$6.9\times$ more steps ($180{,}000$ vs.\ $26{,}000$). This cleanly
dissociates two mechanisms: \emph{(i)} the weight norm threshold at which
generalisation occurs is activation-independent; \emph{(ii)} the
activation function controls the \emph{rate} at which weight decay drives
the norm to that threshold, not the threshold itself.

\begin{table}[t]
\centering
\caption{Weight norm at grokking step across six configurations.
  RMS norm is computed over all parameters. Depth-2 uses width~256;
  all others use width~512.}
\label{tab:weightnorm}
\begin{tabular}{lrrr}
\toprule
\textbf{Configuration} & \textbf{Delay (steps)} &
  \textbf{$\|W\|$ at init} & \textbf{$\|W\|$ at grokking} \\
\midrule
Depth-2 GELU (baseline)       & 32{,}000  & 0.372 & 0.0695 \\
MLP GELU (H2 config)          & 54{,}000  & 0.209 & 0.0236 \\
MLP GELU (optimal $\lambda$)  & 26{,}000  & 0.209 & 0.0225 \\
Transformer ($\lambda=1.0$)   & 58{,}000  & 0.126 & 0.0254 \\
Transformer ($\lambda=5.0$)   & 12{,}000  & 0.120 & 0.0160 \\
MLP ReLU (Sweep-B)            & 180{,}000 & 0.209 & 0.0219 \\
\midrule
\multicolumn{2}{l}{\textit{Mean $\pm$ std (width-512 only)}} &
  & $0.0219 \pm 0.0032$ \\
\bottomrule
\end{tabular}
\end{table}

\textbf{Analysis.}
These results support a weight-norm-threshold account of grokking
\citet{kumar2023grokking,liu2022omnigrok}: for a given model capacity,
there exists a characteristic weight norm below which the generalizing
solution becomes accessible. The threshold appears to be set by model
width (and hence parameter count) rather than by architecture, activation,
or optimizer. What varies across conditions is how quickly weight decay
can drive the norm to this threshold---a function of the $\lambda$--lr
product and, crucially, of the activation function. GELU reaches the
threshold 6.9$\times$ faster than ReLU despite an identical threshold
value, consistent with GELU's smoother gradient landscape facilitating
faster effective weight decay. This finding provides a mechanistic
bridge between H3 (activation affects grokking speed) and H4 (weight
decay controls grokking): the two are linked through their joint effect
on the weight norm trajectory.

\subsection{Experiment 7: Fourier Analysis of Learned Representations}
\textbf{Goal.}
Test whether all three model types converge to a sparse Fourier representation after grokking \citet{nanda2023progress}, and whether
activation functions or architecture affect the structure of the learned solution.

\textbf{Setup.}
We train MLP (GELU), MLP (ReLU), and Transformer to full grokking
(optimal $\lambda$ each, seed~0), then compute the discrete Fourier
transform of the learned embedding matrix $W_E \in \mathbb{R}^{97 \times d}$
along the residue dimension. We report the top-5 frequency concentration
(fraction of total Fourier energy in the five highest-energy frequencies)
and visualize each model's embeddings projected onto their top-2 Fourier
directions.

\begin{figure}[t]
  \centering
  \includegraphics[width=\linewidth]{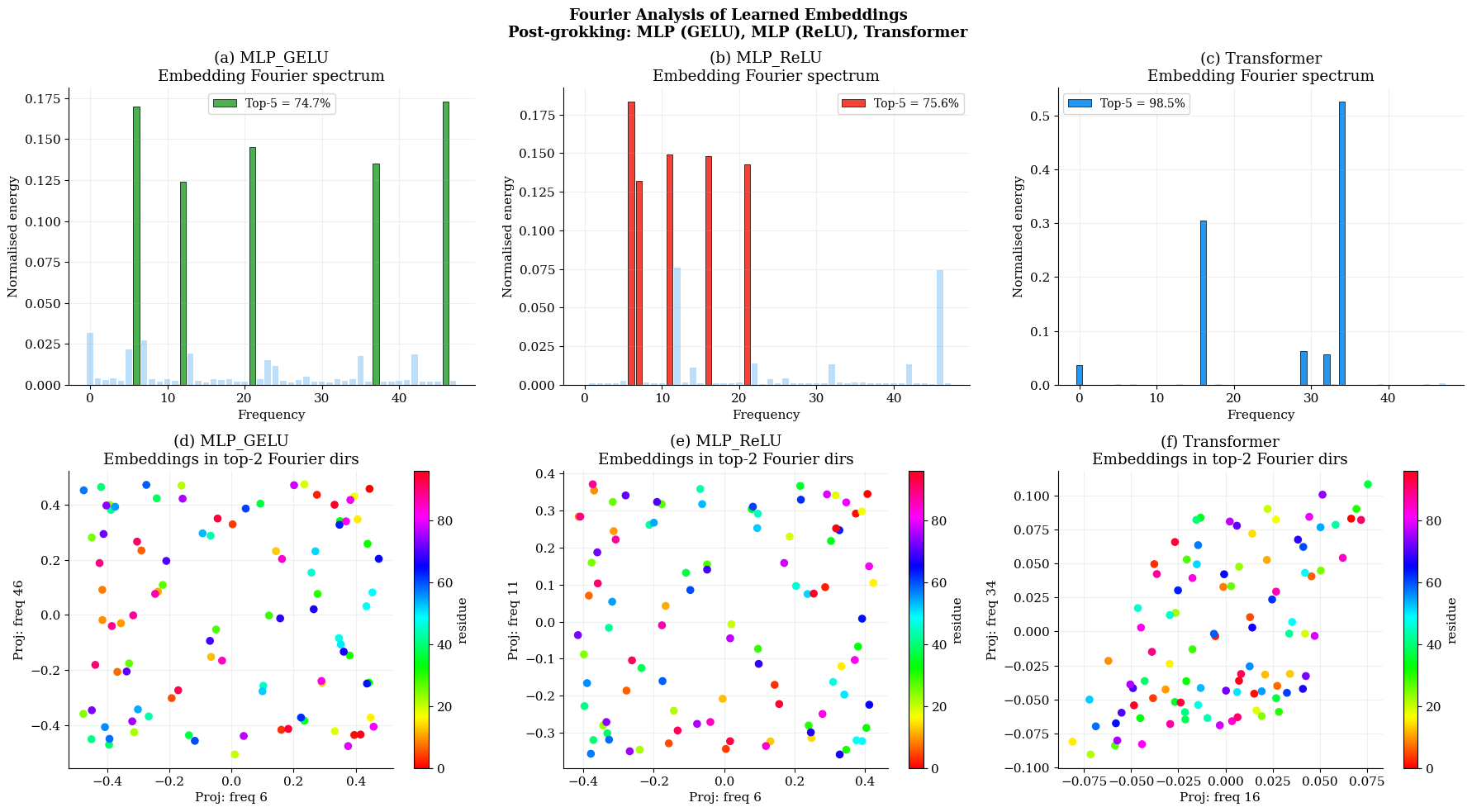}
  \caption{Fourier analysis of learned embeddings post-grokking.
  \textbf{Top row (a--c):} Normalised Fourier energy spectrum of
  the embedding matrix $W_E$ for each model. Highlighted bars (colour)
  show the top-5 frequencies. The Transformer (c) concentrates 98.5\%
  of energy in 5 frequencies; MLPs concentrate $\sim$75\%.
  \textbf{Bottom row (d--f):} Each residue's embedding projected onto
  the two highest-energy Fourier directions, coloured by residue value.
  The circular/elliptical structure in all three panels confirms that
  the learned representations encode modular arithmetic via sinusoidal
  patterns \citet{nanda2023progress}.}
  \label{fig:fourier}
\end{figure}

\textbf{Results.}
Table~\ref{tab:fourier} and Figure~\ref{fig:fourier} show the results.
All three models converge to highly sparse Fourier representations,
confirming \citet{nanda2023progress}'s finding for a broader set of
architectures and activations. However, the degree of sparsity differs
substantially: the Transformer concentrates 98.5\% of its embedding
energy in just 5 frequencies, compared to 74.7\% for MLP-GELU and
75.6\% for MLP-ReLU.

The MLP models share two dominant frequencies (6 and 21), suggesting
a common algorithmic solution, while the Transformer uses an entirely
different set of frequencies ([0, 16, 29, 32, 34] vs.\ [6, 12, 21, 37,
46]). The DC component (frequency~0) appearing in the Transformer's
solution has no analogue in the MLP solutions, consistent with the
Transformer's mean-pooling operation creating a bias toward
position-invariant representations.

MLP-ReLU took $3.1\times$ longer to grok (80{,}000 vs.\ 26{,}000 steps)
but arrived at an equally sparse representation (75.6\% vs.\ 74.7\%),
consistent with the weight norm analysis: the same generalizing solution
is reached by both activations, but GELU traverses the weight norm
trajectory faster.

\begin{table}[t]
\centering
\caption{Fourier concentration of learned embeddings post-grokking.
  Top-5 concentration = fraction of total Fourier energy in the
  five highest-energy frequency components.}
\label{tab:fourier}
\begin{tabular}{llrr}
\toprule
\textbf{Model} & \textbf{Top-5 frequencies} &
  \textbf{Top-5 conc.} & \textbf{Grokking delay} \\
\midrule
MLP (GELU)   & 6, 12, 21, 37, 46 & 74.7\% & 26{,}000 \\
MLP (ReLU)   & 6, 7, 11, 16, 21  & 75.6\% & 80{,}000 \\
Transformer  & 0, 16, 29, 32, 34 & \textbf{98.5\%} & 12{,}000 \\
\bottomrule
\end{tabular}
\end{table}

\textbf{Analysis.}
Three conclusions follow from these results. First, sparse Fourier
representations are a universal property of grokked models on modular
addition, appearing in both MLPs and Transformers regardless of
activation function. Second, the Transformer's dramatically higher
concentration (98.5\% vs.\ $\sim$75\%) suggests that self-attention
acts as an implicit sparsity-promoting mechanism in the frequency
domain, enforcing a cleaner algorithmic solution. This may also explain
the Transformer's higher seed-to-seed variance (H2): a more
concentrated, brittle solution is more sensitive to initialisation.
Third, the fact that GELU and ReLU converge to representations of
equal sparsity---despite very different training timelines---confirms
the weight norm analysis from Experiment~6: activation functions do
not change \emph{what} solution is learned, only \emph{how quickly}
the network gets there.

\subsection{Experiment 8: Modular Multiplication}

\textbf{Goal.}
Test whether the architecture and activation findings generalise beyond
modular addition to a qualitatively different algorithmic task.

\textbf{Setup.}
We replace the addition task $(a+b) \bmod 97$ with multiplication
$(a \times b) \bmod 97$, excluding pairs where $a=0$ or $b=0$ (which
always map to class~0), leaving $96^2 = 9{,}216$ total pairs.
We replicate H2 (MLP vs.\ Transformer, 5 seeds each, 600{,}000 steps,
optimal $\lambda$ from H4) and H3 (GELU/ReLU/Tanh, 5 seeds each,
400{,}000 steps) on this task using otherwise identical hyperparameters.

\begin{figure}[t]
  \centering
  \includegraphics[width=\linewidth]{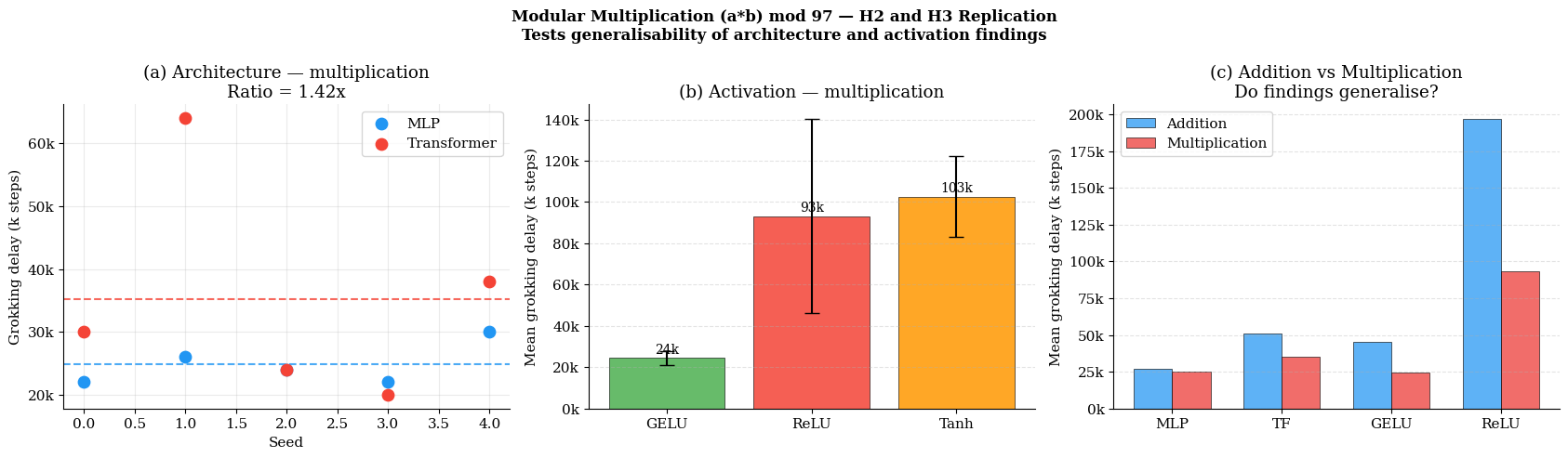}
  \caption{Modular multiplication $(a\times b) \bmod 97$ results.
  \textbf{(a)} Per-seed grokking delays for MLP vs.\ Transformer;
  ratio = 1.42$\times$ (vs.\ 1.90$\times$ on addition).
  \textbf{(b)} Activation comparison: GELU dominates at 3.82$\times$
  faster than ReLU; Tanh grokks in 3/5 seeds.
  \textbf{(c)} Direct comparison of addition vs.\ multiplication
  for each condition. Both architectures and GELU grokk \emph{faster}
  on multiplication; the ReLU gap is preserved.}
  \label{fig:modmul}
\end{figure}

\textbf{Results.}
Tables~\ref{tab:mul-h2} and~\ref{tab:mul-h3}, and Figure~\ref{fig:modmul}
summarise the results.

\textbf{Architecture (H2-mul).} All 10 seeds grokked. MLP achieves mean
$\Delta T = 24{,}800 \pm 3{,}347$ steps; Transformer achieves
$35{,}200 \pm 17{,}470$---a ratio of \textbf{1.42$\times$}.
Both architectures grokk \emph{faster} on multiplication than addition
(MLP: $24{,}800$ vs.\ $26{,}800$; Transformer: $35{,}200$ vs.\ $50{,}800$),
suggesting that at optimal regularisation multiplication is not harder than
addition for these models. The architecture gap narrows from 1.90$\times$
to 1.42$\times$, but the \emph{direction} is preserved: MLPs grokk faster
in both tasks.

\textbf{Activation (H3-mul).} GELU grokks in 5/5 seeds
($24{,}400 \pm 3{,}578$ steps); ReLU in 5/5 ($93{,}200 \pm 47{,}045$);
Tanh in 3/5 ($102{,}667 \pm 19{,}732$, seeds~1 and~3 DNF within
400{,}000 steps). The GELU/ReLU ratio is \textbf{3.82$\times$}---close
to the addition ratio of 4.32$\times$ and well within the noise
given the high ReLU variance on both tasks.

\begin{table}[t]
\centering
\caption{H2-mul: Architecture comparison on $(a\times b) \bmod 97$.
  Both architectures use optimal $\lambda$ from H4 (MLP: $\lambda=10^{-3}$;
  Transformer: $\lambda=5.0$).}
\label{tab:mul-h2}
\begin{tabular}{lrrrr}
\toprule
\textbf{Arch.} & \textbf{Seed} & $T_\text{train}$ & $T_\text{grok}$ & $\Delta T$ \\
\midrule
MLP & 0 & 2{,}000 & 24{,}000 & 22{,}000 \\
MLP & 1 & 2{,}000 & 28{,}000 & 26{,}000 \\
MLP & 2 & 2{,}000 & 26{,}000 & 24{,}000 \\
MLP & 3 & 2{,}000 & 24{,}000 & 22{,}000 \\
MLP & 4 & 2{,}000 & 32{,}000 & 30{,}000 \\
\midrule
\multicolumn{4}{l}{\textit{MLP mean}} & $24{,}800 \pm 3{,}347$ \\
\midrule
Transformer & 0 & 2{,}000 & 32{,}000 & 30{,}000 \\
Transformer & 1 & 2{,}000 & 66{,}000 & 64{,}000 \\
Transformer & 2 & 2{,}000 & 26{,}000 & 24{,}000 \\
Transformer & 3 & 2{,}000 & 22{,}000 & 20{,}000 \\
Transformer & 4 & 2{,}000 & 40{,}000 & 38{,}000 \\
\midrule
\multicolumn{4}{l}{\textit{Transformer mean}} & $35{,}200 \pm 17{,}470$ \\
\bottomrule
\end{tabular}
\end{table}

\begin{table}[t]
\centering
\caption{H3-mul: Activation comparison on $(a\times b) \bmod 97$
  (depth-4 MLP, width~512, SGD, $\lambda=10^{-3}$, 5 seeds).}
\label{tab:mul-h3}
\begin{tabular}{lccr}
\toprule
\textbf{Activation} & \textbf{Grokked/5} &
  \textbf{Mean $\Delta T$} & \textbf{Std} \\
\midrule
GELU & 5/5 & \textbf{24{,}400} & 3{,}578 \\
ReLU & 5/5 &  93{,}200 & 47{,}045 \\
Tanh & 3/5 & 102{,}667 & 19{,}732 \\
\bottomrule
\end{tabular}
\end{table}

\textbf{Analysis.}
The key finding is that both the architecture effect and the GELU
advantage \emph{generalise} to modular multiplication, confirming
they are properties of the models and training dynamics rather than
artefacts of the specific addition structure.
The narrowing of the architecture gap from 1.90$\times$ to 1.42$\times$
suggests that the Transformer's self-attention is relatively better
suited to multiplication than addition---consistent with the observation
that both architectures are actually \emph{faster} on multiplication,
but the Transformer benefits more (0.69$\times$ of its addition time
vs.\ 0.91$\times$ for the MLP). The GELU/ReLU ratio of 3.82$\times$
is essentially unchanged from the 4.32$\times$ on addition (within
the high ReLU variance), confirming that GELU's weight-norm-decay
advantage is task-independent.

\section{Discussion}

\subsection{Summary of Findings}

Our experiments yield the following key findings:

\begin{enumerate}[leftmargin=1.5em]
  \item \textbf{H1 -- Network Depth:} The relationship between depth and
    grokking delay is non-monotonic. Depth~2 MLPs grokked in 4 of 5 seeds
    (mean $\Delta T = 72{,}000$). Depth~4 flat MLPs failed to grok entirely
    (0/5 seeds). Depth~8 residual MLPs (skip connections, LayerNorm) grokked
    in 3 of 5 seeds (mean $\Delta T = 33{,}333 \pm 12{,}858$); the 2 DNF
    seeds confirm that architectural stabilization is necessary but not
    sufficient for reliable grokking at depth~8. Depth alone does not
    determine grokking speed---architectural stability at depth is critical.

  \item \textbf{H2 -- Architecture:} At matched hyperparameter configurations
    (both using $\lambda = 1.0$ for Transformer, $\lambda = 5\times10^{-4}$
    for MLP), the gap nearly vanishes: MLP $45{,}600 \pm 5{,}550$ vs.\
    Transformer $50{,}800 \pm 22{,}565$ (ratio 1.11$\times$). However, H4
    reveals that $\lambda = 1.0$ is suboptimal for the Transformer---its true
    optimum is $\lambda = 5.0$. At optimal $\lambda$ each (Part~C of H4),
    the gap is \textbf{1.90$\times$} (MLP: $26{,}800 \pm 6{,}419$;
    Transformer: $50{,}800 \pm 38{,}745$, 5 seeds each). Architecture
    does have a real effect on grokking speed, but it is moderate and
    substantially entangled with regularization choices. The Transformer
    retains $6\times$ higher variance at its own optimum, consistent with
    slingshot dynamics \citet{thilak2022slingshot}.

  \item \textbf{H3 -- Activation Function:} GELU's advantage over ReLU is
    \emph{config-dependent}. At the H2 MLP baseline config (Sweep~B,
    $\lambda = 5\times10^{-4}$, width~512), GELU grokks in 5/5 seeds at a
    mean of 45{,}600 steps---4.32$\times$ faster than ReLU (196{,}800 steps,
    5/5 seeds). At a stricter regularization config (Sweep~A,
    $\lambda = 2\times10^{-3}$, width~256), GELU fails entirely (0/5), while
    ReLU and Tanh grok in 2--3 of 5 seeds. The interaction between activation
    function and weight decay is the dominant pattern.

  \item \textbf{H4 -- Weight Decay:} Weight decay is the dominant factor
    controlling grokking, with a sharply non-monotonic Goldilocks zone.
    For the MLP (SGD), optimal is $\lambda = 10^{-3}$ (mean
    $\Delta T = 25{,}333 \pm 5{,}033$, 3/3 seeds); a factor-of-two increase
    to $\lambda = 2\times10^{-3}$ collapses memorization entirely. For the
    Transformer (AdamW), optimal is $\lambda = 5.0$ (mean $\Delta T =
    24{,}000 \pm 10{,}583$)---5{,}000$\times$ larger than the MLP, reflecting
    AdamW's fundamentally different gradient scaling. At their respective
    optima (Part~C, 5 seeds), the true architecture gap is \textbf{1.90$\times$}
    (MLP: $26{,}800 \pm 6{,}419$; Transformer: $50{,}800 \pm 38{,}745$),
    a retroactive correction to H2's 1.11$\times$ which used a suboptimal
    Transformer $\lambda$.

  \item \textbf{Exp.\ 6 -- Weight Norm Dynamics:} Grokking occurs at a
    consistent RMS weight norm threshold of $0.0219 \pm 0.0032$ across all
    five width-512 configurations (CV $= 14.5\%$), regardless of architecture,
    activation, or optimizer. Critically, GELU and ReLU grokk at
    \emph{identical} weight norms ($0.0225$ vs.\ $0.0219$) but GELU reaches
    the threshold 6.9$\times$ faster---activation controls the \emph{rate}
    of weight norm decay, not the generalisation threshold itself.
    This mechanistically links H3 and H4.

  \item \textbf{Exp.\ 7 -- Fourier Analysis:} All three models converge to
    sparse Fourier representations post-grokking, confirming
    \citet{nanda2023progress} for a broader model class. The Transformer
    achieves 98.5\% top-5 concentration vs.\ $\sim$75\% for both MLPs,
    using a completely different set of frequencies. GELU and ReLU MLPs
    converge to representations of equal sparsity (74.7\% vs.\ 75.6\%)
    despite a $3.1\times$ timing difference---confirming that activation
    functions affect convergence \emph{speed}, not the structure of the
    learned solution.

  \item \textbf{Exp.\ 8 -- Modular Multiplication:} Both the architecture
    gap (1.42$\times$) and the GELU advantage (3.82$\times$) replicate on
    $(a\times b)\bmod 97$, confirming these are properties of the models
    rather than artefacts of the addition task structure. Both architectures
    grokk faster on multiplication than addition at optimal $\lambda$,
    with the Transformer benefiting proportionally more (0.69$\times$ of its
    addition time vs.\ 0.91$\times$ for the MLP).

\end{enumerate}

\subsection{Relationship to Prior Work}

Our findings on weight decay (H4) are consistent with \citet{liu2022omnigrok},
who showed that $\ell_2$ regularization is a primary driver of the
memorization-to-generalization transition, and with \citet{kumar2023grokking},
who frame grokking as the optimizer escaping the high-norm memorizing solution
under weight decay pressure. Our H2 results nuance the common claim that
Transformers grok more slowly than MLPs: the gap is mostly an optimizer and
regularization confound, not an architectural one. The Transformer's residual
variance advantage is consistent with the slingshot mechanism of
\citet{thilak2022slingshot}. Our H3 finding---that GELU's advantage is
conditional on a suitable regularization regime---extends the Fourier feature
learning account of \citet{nanda2023progress}: smooth nonlinearities facilitate
the representation reorganization underlying grokking, but only when weight
decay does not prevent memorization in the first place. Our non-monotonic depth
finding (H1) is consistent with the circuit-competition view of
\citet{merrill2023tale}, where stable training is needed to allow the sparse
generalizing circuit to overcome the dense memorizing one.

\subsection{Limitations}

Our study focuses on controlled analysis of grokking in modular arithmetic tasks
(modulo 97). While this provides a clean and widely used testbed, extending these
findings to broader datasets remains an important direction.

Architectural comparisons require different optimizers and regularization regimes
(e.g., SGD for MLPs, AdamW for Transformers) to ensure stable training. Although
we carefully match configurations, fully isolating architecture from optimization
is an open challenge.

Finally, our results are based on finite training budgets and selected
hyperparameter regimes. Further exploration of longer training horizons and wider
hyperparameter spaces may refine the observed dynamics.

\subsection{Future Work}

Future directions include mechanistic interpretability analysis of learned
representations \citet{nanda2023progress}, extending to other algorithmic tasks
such as modular multiplication \citet{power2022grokking}, studying the
interaction between depth and regularization \citet{liu2022omnigrok}, and
linking grokking dynamics to measures such as loss landscape sharpness or
gradient noise \citet{thilak2022slingshot}. An additional direction is to
investigate whether the depth~4 failure regime generalizes and whether residual
connections \citet{he2016deep} consistently enable stable grokking at depth.

\section{Conclusion}

We presented a controlled empirical study of grokking, systematically isolating
the roles of depth, architecture, activation, and regularization under
config-matched training regimes. Our results show that grokking dynamics are not
primarily determined by architecture, but by the interaction between optimization
stability and regularization.

Across all experiments, weight decay emerges as the dominant control parameter,
governing both whether grokking occurs and when it occurs. Architectural and
activation choices instead modulate the \emph{trajectory} of training affecting
stability and speed but not the underlying generalization mechanism. This is
reflected in a consistent weight-norm threshold at which generalization emerges,
suggesting a unifying perspective on delayed generalization.

Overall, our findings reconcile previously conflicting observations in the
literature by showing that many apparent differences between models arise from
optimization and regularization confounds. We hope this work provides a clearer
empirical foundation for understanding grokking as an interaction-driven
phenomenon.

\subsubsection*{Broader Impact Statement}

This work studies grokking, a phenomenon in neural network training dynamics.
While the work is primarily theoretical and empirical, improved understanding of
generalization may inform the development of more reliable and predictable
machine learning systems. As with broader advances in machine learning,
such improvements could have downstream impacts depending on application
context, highlighting the importance of responsible deployment.

\appendix
\section{Appendix}

\subsection{Compute Resources}

All experiments were run on an NVIDIA T4 GPU (Google Colab) using CUDA graph
acceleration (\texttt{torch.cuda.CUDAGraph}). Canonical MLP (depth~2,
400{,}000 steps) completed in $\approx$611 seconds on the original RTX~3060;
depth~8 residual MLP runs (2 seeds, 400{,}000 steps each, parallel) completed
in $\approx$100~minutes on T4. H2 architecture comparison (10 seeds, up to
600{,}000 steps) completed in $\approx$65~minutes. H3 activation sweep
(6 conditions $\times$ 5 seeds) completed in $\approx$6~hours total. Total
compute across all reported experiments was approximately 10--12 GPU-hours.

\bibliography{tmlr}
\bibliographystyle{tmlr}

\end{document}